\documentclass[twoside,11pt,abbrvbib]{article}

\usepackage{blindtext}

\usepackage[preprint]{jmlr2e}
\usepackage{listings}
\usepackage{xcolor}
\PassOptionsToPackage{hyphens}{url}\usepackage{hyperref}

\definecolor{codegreen}{rgb}{0,0.6,0}
\definecolor{codegray}{rgb}{0.5,0.5,0.5}
\definecolor{codepurple}{rgb}{0.58,0,0.82}
\definecolor{backcolour}{rgb}{0.97,0.97,0.97}

\lstdefinestyle{python}{
    backgroundcolor=\color{backcolour},
    commentstyle=\color{codegreen},
    keywordstyle=\color{blue},
    numberstyle=\tiny\color{codegray},
    stringstyle=\color{codepurple},
    basicstyle=\ttfamily\footnotesize,
    breakatwhitespace=false,
    breaklines=true,
    captionpos=b,
    keepspaces=true,
    numbers=left,
    numbersep=5pt,
    showspaces=false,
    showstringspaces=false,
    showtabs=false,
    tabsize=2
}

\usepackage{lastpage}
\jmlrheading{0}{0}{1-\pageref{LastPage}}{0; Revised 0}{0}{0}{Matthew Middlehurst, Ali Ismail-Fawaz, Antoine Guillaume, Christopher Holder, David Guijo-Rubio, Guzal Bulatova, Leonidas Tsaprounis, Lukasz Mentel, Martin Walter, Patrick Sch\"afer, Anthony Bagnall}

\ShortHeadings{aeon: a Python toolkit for learning from time series}{Middlehurst \textit{et al.}}
\firstpageno{1}

\begin{document}

\title{aeon: a Python toolkit for learning from time series}

\author{\name Matthew Middlehurst\textsuperscript{1} \email m.b.middlehurst@soton.ac.uk \\
    \name Ali Ismail-Fawaz\textsuperscript{2} \email ali-el-hadi.ismail-fawaz@uha.fr \\
    \name Antoine Guillaume\textsuperscript{3} \email antoine.guillaume@ensta-paris.fr \\
    \name Christopher Holder\textsuperscript{4} \email c.holder@uea.ac.uk \\
    \name David Guijo-Rubio\textsuperscript{4,5} \email dguijo@uco.es \\
    \name Guzal Bulatova \email guzalbulatova@gmail.com \\
    \name Leonidas Tsaprounis \email leonidas.tsap@gmail.com \\
    \name Lukasz Mentel \email lukasz.mentel@pm.me \\
    \name Martin Walter \email martin.friedrich.walter@gmail.com \\
    \name Patrick Sch\"afer\textsuperscript{6} \email \email patrick.schaefer@hu-berlin.de \\
    \name Anthony Bagnall\textsuperscript{1} \email a.j.bagnall@soton.ac.uk \\  
    \textit{The authors are the developers of aeon at the time of submission, with the following affiliations:} \\
    \textit{\textsuperscript{1}ECS, University of Southampton, United Kingdom}
    \textit{\textsuperscript{2}IRIMAS, Université de Haute-Alsace, France}
    \textit{\textsuperscript{3}ENSTA Paris, France}
    \textit{\textsuperscript{4}CMP, University of East Anglia, United Kingdom}
    \textit{\textsuperscript{5}DIAN, University of Córdoba, Spain}
    \textit{\textsuperscript{6}Humboldt-Universit\"at zu Berlin, Germany}
    }

\editor{TODO}

\maketitle

\begin{abstract}%
\texttt{aeon} is a unified Python $3$ library for all machine learning tasks involving time series. The package contains modules for time series forecasting, classification, extrinsic regression and clustering, as well as a variety of utilities, transformations and distance measures designed for time series data. \texttt{aeon} also has a number of experimental modules for tasks such as anomaly detection, similarity search and segmentation. \texttt{aeon} follows the \texttt{scikit-learn} API as much as possible to help new users and enable easy integration of \texttt{aeon} estimators with useful tools such as model selection and pipelines. It provides a broad library of time series algorithms, including efficient implementations of the very latest advances in research. Using a system of optional dependencies, \texttt{aeon} integrates a wide variety of packages into a single interface while keeping the core framework with minimal dependencies. The package is distributed under the $3$-Clause BSD license and is available at \url{https://github.com/aeon-toolkit/aeon}.

This version was submitted to the JMLR journal on 02 Nov 2023 for v0.5.0 of \texttt{aeon}. At the time of this preprint \texttt{aeon} has released v0.9.0, and has had substantial changes.
\end{abstract}



\section{Introduction}
\lstset{style=python}

Time series appear in all areas of scientific research and play a central role in all business analytics. 
Time Series Machine Learning (TSML) research involves developing and using algorithms that exploit the unique characteristics of ordered data: interesting features may be based on the interaction between observations. TSML encompasses standard machine learning tasks such as classification, clustering, extrinsic regression and anomaly detection in addition to time series specific problems such as forecasting and segmentation. TSML is an active research field, and \texttt{aeon} plays a central role in facilitating reproducible research~\citep{middlehurst2024bake,guijo2024unsupervised,holder2023review}.

\texttt{aeon} is a Python toolkit for TSML tasks, preprocessing and benchmarking. The package is designed to be easy to use, especially for those familiar with \texttt{scikit-learn}~\citep{pedregosa2011scikit}. Where possible the \texttt{scikit-learn} API and style is followed, and where there are shared learning tasks \texttt{aeon} objects aim to be compatible with the available utilities such as model selection and 
pipelining. \texttt{aeon} is forked from version v$0.16.0$ of the \texttt{sktime}~\citep{loning2019sktime} package. Since the community split, the packages have diverged significantly in terms of available modules and included estimators. In the following, we provide an overview of the packages available in \texttt{aeon}, as well as some experimental and upcoming modules. This document describes version v$0.5.0$ of \texttt{aeon}. \texttt{aeon} supports all versions from Python $3.8$ onwards in this version, and online documentation is available at \url{https://aeon-toolkit.org}. 

\section{Code Design and Implementation}

In \texttt{aeon}, the package design is modular, with algorithms grouped by learning tasks. The primary TSML modules such as forecasting and classification are encapsulated as much as possible and do not import from each other. Supporting modules such as distances and transformations exist to hold functions and classes useable in multiple learning tasks, as well as standalone if required.

\begin{figure}
    \centering
    \includegraphics[width=0.9\linewidth,trim={2cm 5cm 2cm 4.5cm},clip]{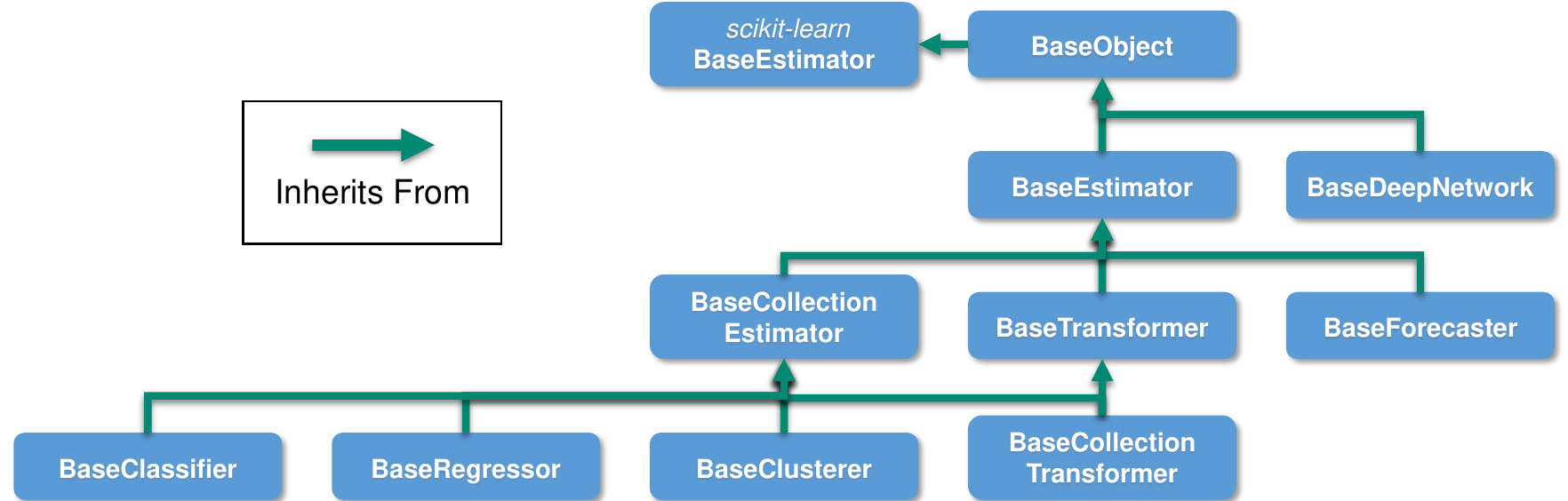}
    \caption{A flowchart of base class inheritance for the main \texttt{aeon} modules. Classification algorithms will inherit from \texttt{BaseClassifier}, for example.}
    \label{fig:class-flowchart}
\end{figure}

\texttt{aeon} uses object-oriented design for most cases to fit into the \texttt{scikit-learn} estimator interface. Some modules are more functional in design, such as distance measures and performance metrics. \texttt{aeon} TSML classes follow an inheritance structure with each module having its own class. An example of this structure for some of the main \texttt{aeon} modules is shown in Figure~\ref{fig:class-flowchart}. The base class for a module contains mandatory methods such as \texttt{fit} to be inherited, as well default functionality such as converting and verifying input data.

Estimators have tags that identify the types of data the estimators can take and their functionality. So, for example, a classifier with the following capability tags can handle both multivariate and unequal length time series input. 
\begin{lstlisting}[language=Python]
_tags = {
    "capability:multivariate": True,
    "capability:unequal_length": True
}
\end{lstlisting}

In \texttt{aeon} we aim to keep the package core dependencies to a minimum. The primary dependency is \texttt{scikit-learn}, which we base our estimator interface off. Shared with \texttt{scikit-learn}, the package contains extensive usage of the \texttt{numpy}~\citep{harris2020array} and \texttt{scipy}~\citep{virtanen2020scipy} libraries. While \texttt{scikit-learn} has \texttt{pandas}~\citep{mckinney2010data} as an optional dependency, it is required in \texttt{aeon} and has extensive usages in modules such as forecasting. \texttt{scikit-learn} makes use of \texttt{cython}~\citep{behnel2010cython} to optimise its implementation, \texttt{aeon} depends on \texttt{numba}~\citep{lam2015numba} and aims to use \texttt{numba} compiled just-in-time (JIT) functions where possible.

As well as the core dependencies, \texttt{aeon} also includes a range of optional/soft dependencies to packages such as \texttt{statsmodels}~\citep{seabold2010statsmodels}, \texttt{tensorflow}~\citep{tensorflow2015-whitepaper}, and \texttt{tsfresh}~\citep{christ2018time}. These are commonly used to create wrappers for algorithms present in these packages or used as a framework for estimators such as deep learners. Creating a singular framework for these TSML algorithms allows for easier benchmarking and reproducibility.

\section{Time Series Modules}

\begin{figure}
    \centering
    \includegraphics[width=0.9\linewidth,trim={0cm 0.25cm 0cm 0.5cm},clip]{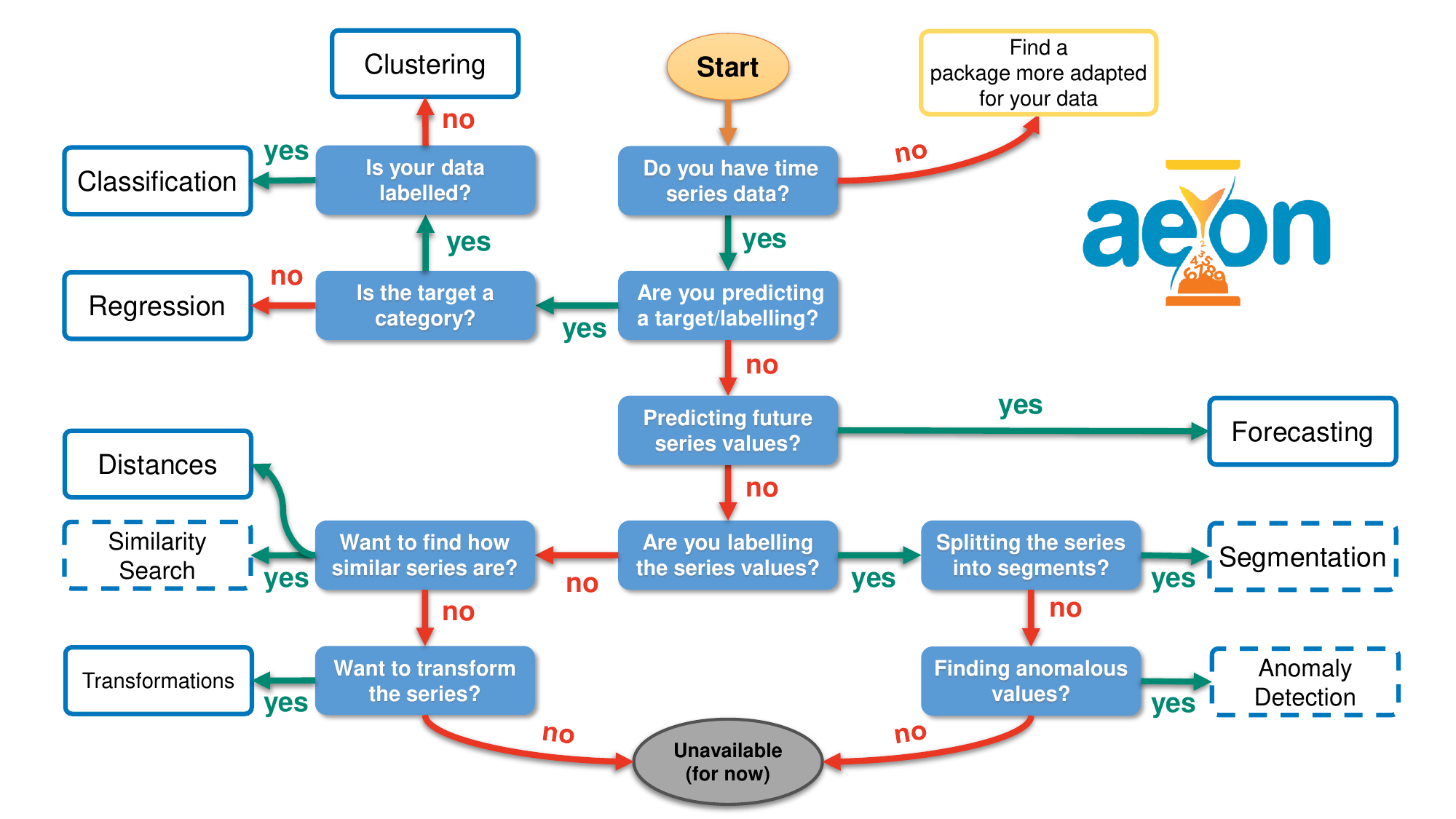}
    \caption{A flowchart for selecting the correct time series learning module in \texttt{aeon}. Learning tasks with dashed borders are experimental.}
    \label{fig:module-flowchart}
\end{figure}

\texttt{aeon} splits different TSML tasks into modules. In the following we go over some of the stable core modules of \texttt{aeon}, as well as some experimental modules. Figure~\ref{fig:module-flowchart} displays a flowchart for different TSML modules, with situations where one would want to use each.

\subsection{Forecasting}

At its simplest, forecasting involves predicting the next values in sequence of a given time series. 
\texttt{aeon} provides a unified interface for a range of popular forecasting tools and models, including \texttt{prophet}~\citep{taylor2018forecasting}, \texttt{pmdarima}, \texttt{statsforecast}, \texttt{statsmodels}~\citep{seabold2010statsmodels} and \texttt{tbats}~\citep{de2011forecasting}.

There are mechanisms for reducing forecasting to regression through windowing and a range of tools to help produce better forecasts, including: model selection, evaluation, composition, ensembling; probabilistic/conformal intervals~\citep{stankeviciute2021conformal}, hierarchical forecasting, global forecasting and panel forecasting.

The forecasting interface is built using a similar \textit{fit} and \textit{predict} structure to \texttt{scikit-learn}, used in conjunction with a forecasting horizon. Forecasting works primarily with \texttt{pandas} \texttt{DataFrame} and \texttt{Series} objects, although a range of conversion tools for other data types are available. The basic usage is as follows.

\begin{lstlisting}[language=Python]
from aeon.datasets import load_airline
from aeon.forecasting.trend import TrendForecaster
y = load_airline()
forecaster = TrendForecaster()
forecaster.fit(y)  # fit the forecaster
pred = forecaster.predict(fh=[1,2,3])  # predict the next 3 values
\end{lstlisting}

\subsection{Classification, Clustering and Regression}

Time series classification, clustering and extrinsic regression tasks all have the same \textit{fit} and \textit{predict} interface and are compatible with the \texttt{scikit-learn} interface for the same tasks. All of these estimators inherit from a \texttt{BaseCollectionEstimator}, where tags are defined and conversions are performed. Typically, collections of series are stored in a $3$D \texttt{numpy} array, while unequal length series are stored in a list of $2$D \texttt{numpy} arrays. The shape for both of these data types is (n\_cases, n\_channels, n\_timepoints). Using classification as an example, the most basic usage is as follows:

\begin{lstlisting}[language=Python]
from aeon.datasets import load_classification
from aeon.classification.convolution_based import RocketClassifier
train_X, train_y, _ = load_classification("GunPoint", split="Train")
test_X, _, _ = load_classification("GunPoint", split="Test")
clf = RocketClassifier()
clf.fit(train_X,train_y)  # fit the classifier
preds = clf.predict(test_X)  # make a prediction for each test case
\end{lstlisting}

\subsubsection{Classification}

Classification algorithms are sometimes grouped on the basis of the transformation used~\citep{bagnall2017great,middlehurst2024bake}. \texttt{aeon} contains an extensive range of classifiers, using such a taxonomy to create sub-packages of approaches. 
The classification module includes a wide breadth of algorithmic approaches types: convolution based~\citep{dempster2020rocket,tan2022multirocket}; deep learning~\citep{fawaz2019deep,fawaz2020inceptiontime,ismail2022deep}; dictionary based~\citep{schafer2015boss,middlehurst2020temporal,schafer2023weasel,bennett2023red}; distance based~\citep{lines2015time,lucas2019proximity}; feature based~\citep{lubba2019catch22,middlehurst2022freshprince}; hybrids~\citep{lines2018time,middlehurst2021hive}; interval based~\citep{deng2013time,middlehurst2020canonical,cabello2023fast}; and shapelet based~\citep{bostrom2017binary,nguyen2021mrsqm,guillaume2022random}.
Our documentation and the research field survey in \cite{middlehurst2024bake} has more information on each category and algorithm.

\subsubsection{Clustering}

Clustering works in conjunction with the distances module to provide $k$-means based and $k$-medoids based clustering. $k$-means works with over ten elastic distance functions implemented~\citep{holder2023review} and can be used with barycentre averaging~\citep{petitjean2011global,ismail2023shapedba}. We are also working on including a range of feature based and deep learning clustering algorithms~\citep{lafabregue2022end}. 

\subsubsection{Regression}

Extrinsic regression is a recently defined task~\citep{tan2021time}. \texttt{aeon} contains a range of algorithms adapted from classification~\citep{guijo2024unsupervised,middlehurst2023extracting}, including popular deep learning algorithms~\citep{foumani2023deep}. We hope that keeping open implementations in \texttt{aeon} will help the field in growing, and we intend to keep the library up to date with the latest state-of-the-art.

\subsection{Transformations}

Transformers are objects that transform data from one representation to another. \texttt{aeon} contains time series specific transformers which can be used in pipelines in conjunction with other estimators. \texttt{aeon} transformers follow the \texttt{scikit-learn} design with a \textit{fit} and \textit{transform} method. 

General transformers aim to accept all input types whether that is a single series or a collection, and will attempt to restructure the data or broadcast to multiple transformer objects if necessary to fit the input data to the data structure used by the transformer. Collection transformers are structured to efficiently process collections of series, and can only take that kind of input. Transformations are used extensively in all other modules. 

Transformers come in multiple types. They can be series-to-series transformations which both take and output a time series, such as the Fourier transform or channel selection for multivariate series~\citep{dhariyal2023scalable}. Alternatively, transformers can be series-to-features which take a series input but output a feature vector such as basic summary statistics or TSFresh~\citep{christ2018time}.

\subsection{Experimental Modules}

An experimental module is an area of the code base that in progress of being developed at the time of writing and can still rapidly change. Current experimental modules include: Segmentation~\citep{hallac2019greedy,ermshaus2023clasp}; Anomaly detection~\citep{nakamura2020merlin,talagala2021anomaly}; Similarity search; and benchmarking~\citep{ismail2023approach}.

\section{Conclusions}

There are several other Python toolkits that implement TSML algorithms. These include \texttt{tslearn}~\citep{tavenard2020tslearn} and \texttt{pyts}~\citep{faouzi2020pyts} for classification and clustering, \texttt{adtk} for anomaly detection, numerous time series distance packages such as \texttt{dtaidistance} and matrix profile tools such as \texttt{stumpy}~\citep{law2019stumpy}. We believe \texttt{aeon} is the most comprehensive toolkit for time series machine learning. We hope that our attempt at unifying diverse research fields and communities such as forecasting, classification and anomaly detection will benefit the development of the Python time series ecosystem. \texttt{aeon} joined numFOCUS as an affiliate project in 2024 and we hope to work towards full membership in the near future.

\acks{ 
    \texttt{aeon} is supported by the EPSRC under a special call grant EP/W030756/1. Thank you to everyone who has contributed to \texttt{aeon} whether that be through code, documentation or other means. There are too many to credit individually, but everyone has a hand in making the project great.
}

\bibliography{AnomalyDetection,Classification,Clustering,Forecasting,Regression,Segmentation,Software,Statistics,Transformations,main}

\end{document}